\documentclass[10pt,conference]{IEEEtran}
\usepackage{amsmath,amsfonts}
\usepackage{algorithmic}
\usepackage{algorithm}
\usepackage{array}
\usepackage[caption=false,font=normalsize,labelfont=sf,textfont=sf]{subfig}
\usepackage{stfloats}
\usepackage{url}
\usepackage{verbatim}
\usepackage{graphicx}
\usepackage{amsmath,amssymb,amsfonts}
\usepackage{algorithmic}
\usepackage{graphicx}
\usepackage{textcomp}
\usepackage{xcolor}
\usepackage{csvsimple}
\usepackage{cite}
\usepackage[absolute,overlay]{textpos}
\usepackage{blindtext}
\usepackage{threeparttable}

\usepackage{tikz}
\usetikzlibrary{calc}

\newcommand{\smat}[1]{\left[#1 \times\right]}

\definecolor{bblue}{HTML}{025497}
\definecolor{rred}{HTML}{C0504D}
\definecolor{ggreen}{HTML}{9BBB59}
\definecolor{ppurple}{HTML}{9F4C7C}
\definecolor{ggray}{HTML}{4D4D4D}
\IEEEoverridecommandlockouts

\begin{document}
\title{Integration of Visual SLAM into Consumer-Grade Automotive Localization}

\author{Luis Diener, Jens Kalkkuh and Markus Enzweiler
\thanks{L. Diener and J. Kalkkuhl are with Mercedes-Benz AG, Germany. (luis.diener@mercedes.benz.com)}
\thanks{M. Enzweiler is with the Institute for Intelligent Systems, Esslingen University of Applied Sciences, Germany.}
\thanks{This work is a result of the joint research project STADT:up (19A22006O). The project is supported by the German Federal Ministry for Economic Affairs and Climate Action, based on a decision of the German Parliament. The authors are solely responsible for the content of this publication.}
\thanks{This manuscript has been submitted to the IEEE for possible publication.}
}




\maketitle
\thispagestyle{plain}
\pagestyle{plain}

\begin{abstract}
Accurate ego-motion estimation in consumer-grade vehicles currently relies on proprioceptive sensors, i.e. wheel odometry and IMUs, whose performance is limited by systematic errors and calibration. While visual–inertial SLAM has become a standard in robotics, its integration into automotive ego-motion estimation remains largely unexplored. This paper investigates how visual SLAM can be integrated into consumer-grade vehicle localization systems to improve performance. We propose a framework that fuses visual SLAM with a lateral vehicle dynamics model to achieve online gyroscope calibration under realistic driving conditions. Experimental results demonstrate that vision-based integration significantly improves gyroscope calibration accuracy and thus enhances overall localization performance, highlighting a promising path toward higher automotive localization accuracy. We provide results on both proprietary and public datasets, showing improved performance and superior localization accuracy on a public benchmark compared to state-of-the-art methods.
\end{abstract}


\section{Introduction}
\IEEEPARstart{L}{ocalization} and accurate ego-motion estimation are fundamental for modern driver assistance and automated driving functions. The vehicle's ego-motion state describes its motion relative to an earth-fixed reference frame, including attitude relative to the ground plane, velocity vector, and position. In consumer-grade vehicles, relative localization is typically based on proprioceptive sensors such as wheel odometry, gyroscopes, and accelerometers \cite{Marco.2020}. Although these methods are well established, their performance strongly depends on sensor calibration and vehicle-model assumptions. Automotive-grade proprioceptive sensors provide high levels of integrity and robustness but are prone to systematic errors and insufficient calibration \cite{Marco.2020}. For inertial measurement units (IMUs), online calibration remains particularly challenging, as uncorrected errors rapidly cause drift in the estimated position. 

\begin{figure}[t]
\includegraphics[width=\linewidth]{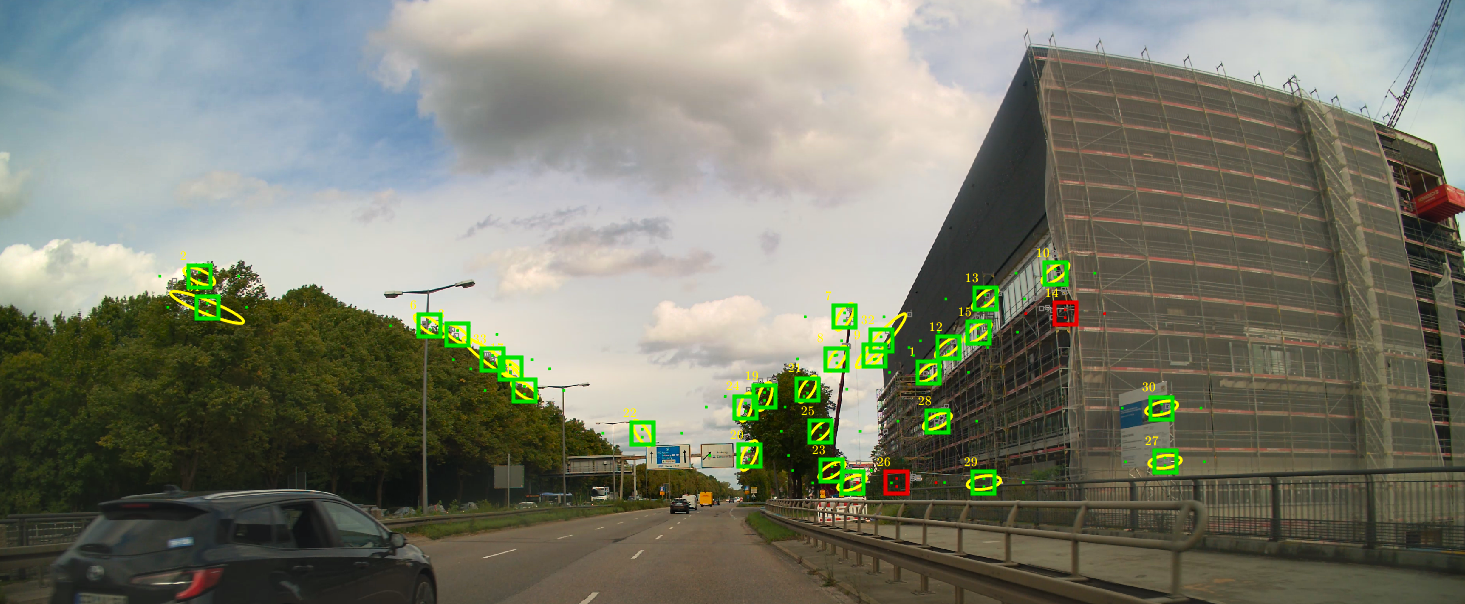}
\caption{On-board camera tracking features in an urban environment. Green indicates successful tracking, with the yellow ellipses showing the uncertainty of the track.}
\label{fig:algo} 
\end{figure}

With the increasing integration of cameras into vehicles, researchers and manufacturers have explored their use for localization \cite{Rabe.2010b, Yang.2024, Liang.2022}. Fig.~\ref{fig:algo} depicts such a front-facing camera, tracking features for localization. While cameras provide valuable information, they remain sensitive to environmental factors such as lighting, weather, and occlusion \cite{Zhang.2023b}. Consequently, they cannot yet satisfy the integrity and robustness requirements of consumer-grade applications, where localization must remain reliable at all times.

In robotics, visual-inertial simultaneous localization and mapping (SLAM) has emerged as a standard localization technique, fusing camera-based tracking with IMU measurements to create a map of the environment \cite{Civera.2008}. While the integration of various proprioceptive sensors and vehicle models into SLAM frameworks has been studied successfully \cite{Lee.2020, Lee.2022, Ma.2019, Li.2024, Vial.2022, Xiong.2022}, the converse integration of visual SLAM into proprioceptive localization algorithms remains largely unaddressed. This raises the question of how much consumer-grade localization systems could benefit from such an integration. One promising application is inertial-sensor calibration, in particular the calibration of the gyroscope. Accordingly, we integrate visual SLAM into a proprioceptive consumer-grade localization stack to perform online gyroscope calibration without vehicle-specific calibration runs. 
Our contributions are:
\begin{itemize}
\item Extending a visual–inertial SLAM calibration framework with a single-parameter lateral-velocity model for consumer-grade applications.
\item Demonstrating online gyroscope calibration under realistic driving conditions using a reduced error model.
\item Designing an adaptive Kalman filter that separates state and parameter estimation for stable online calibration.
\end{itemize}

Through this integration, we aim to bridge the gap between visual SLAM research and real-world automotive applications, offering a path toward more accurate localization in consumer-grade vehicles. Our specific choice to  integrate visual SLAM through gyroscope calibration also addresses a key robustness issue, since the improved proprioceptive localization performance remains consistent even when visual SLAM is temporarily unavailable. Fig.~\ref{fig:method} provides a top-level description of our proposed approach.

In this paper we first discuss the relevant related work in Section II, after which we describe the proposed system in Section III and apply it to the adaptive Kalman filter in Section IV. Section V provides the experimental results with the subsequent conclusions in Section VI.

\section{Related Work}
\subsection{Visual-Inertial Navigation}
Recent progress in SLAM has been driven largely by visual and visual-inertial systems, with algorithms typically divided into batch-optimization and filter-based methods. While optimization approaches such as ORB-SLAM \cite{Campos.2021} and OKVIS \cite{Leutenegger.2015} achieve high accuracy, they demand significant computational resources. Filter-based methods instead trade accuracy for efficiency. There, research has focused on improving consistency through alternative feature parameterizations, ranging from depth–bearing separation \cite{Civera.2008} to robocentric representations \cite{Huai.2018} and minimal rotation parameterizations \cite{Bloesch.2017}.

\begin{figure*}[t]
\centering
\includegraphics[width=\textwidth]{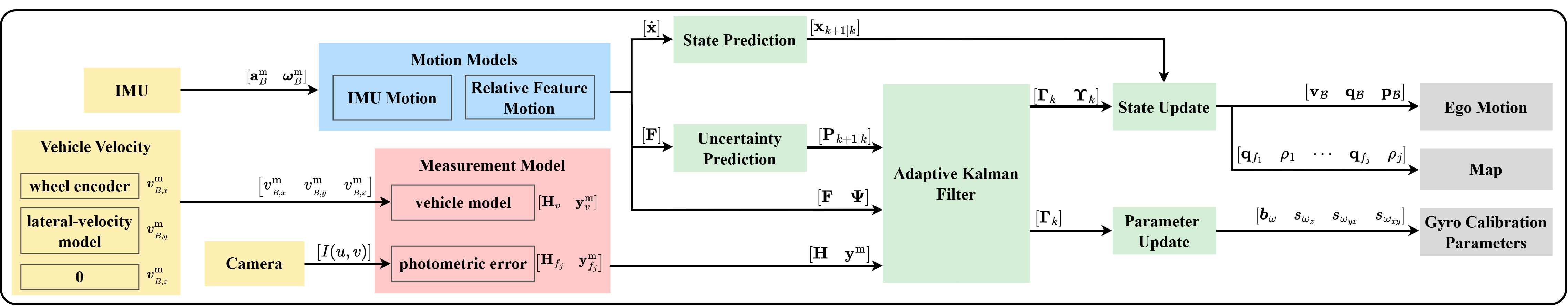}
\caption{Overview of the proposed method. The IMU measurements are used to propagate both the vehicle motion and the relative feature motion. The vehicle's velocity measurements and the camera images are fed through the measurement model. The vehicle-velocity model also includes our lateral-velocity model. Within the adaptive Kalman filter framework the state and uncertainty are predicted and updated accordingly. The feedback loops are omitted for visual clarity.}
\label{fig:method}
\end{figure*}

To enhance robustness, researchers have gradually extended SLAM with additional sensors and motion models, including wheel odometry \cite{Lee.2020}, GNSS \cite{Lee.2022}, and kinematic \cite{Ma.2019, Li.2024} or dynamic vehicle models \cite{Vial.2022, Xiong.2022}. The common pattern in these works is to start with a SLAM framework and then add sensor modalities or model-based constraints. However, in automotive contexts ego-motion estimation has traditionally been built around proprioceptive methods \cite{Marco.2020}. Gentil et al. \cite{Gentil.2025} compared proprioceptive localization with various radar and lidar-based SLAM algorithms. They found that under ideal conditions (no wheel slip and proper calibration) basic proprioceptive localization algorithms can outperform even state-of-the-art SLAM methods. This suggests that the most effective role of SLAM in such systems is not to directly improve localization, but sensor calibration instead. This is especially important in consumer vehicles, which often suffer from poorly calibrated inertial sensors and uncertain vehicle parameters.

\subsection{Automotive IMU Calibration}
A major limitation of automotive localization is insufficient sensor calibration, particularly for consumer-grade IMUs. Calibration methods using vehicle dynamics can, in principle, identify accelerometer and gyroscope errors \cite{Huttner.2018, Marco.2020, Marco.2020b}, but they require high dynamic excitation and are then only weakly observable and inaccurate. GNSS-aided yaw-rate calibration works well in open-sky environments, where accurate heading estimates are available \cite{Mori.2025, Chen.2021}. In urban settings, however, GNSS suffers from limited availability, integrity, and performance. \cite{Obst.2012, Khana.2018}. To address these limitations, cameras provide complementary information unaffected by these issues. Independently of sensor modalities, consumer-grade vehicles can perform stand-still calibration of gyroscope offsets, although the calibration accuracy strongly depends on the frequency and duration of stand-still events.

IMU calibration is also a key focus of research in visual-inertial SLAM \cite{Yang.2024,Yang.2023,Zhao.2023}. Observability analyses \cite{Yang.2020} show that calibration is unfeasible under degenerate motions, such as planar motion or one-axis rotation, which are common in automotive scenarios. Attempts to mitigate this by selecting information-rich motion segments \cite{Schneider.2019} allow for better performance, but accelerometer calibration still requires substantial excitation, similar to purely proprioceptive approaches. This suggests that SLAM is not suitable to improve accelerometer calibration in consumer-grade vehicles. By contrast, gyroscope calibration is less dependent on excitation and can therefore benefit more directly from visual information.

\subsection{Research Gap}
Although visual-inertial SLAM has advanced considerably, most existing research emphasizes extending SLAM frameworks with additional proprioceptive sensors or vehicle models, rather than adapting visual SLAM for consumer-grade automotive localization.
At the same time, a major limitation in consumer-grade localization remains the poor calibration of inertial sensors, particularly of the gyroscope. {This motivates our work:} we investigate how visual SLAM can be integrated into automotive localization not as a replacement, but as a means to improve gyroscope calibration in consumer-grade vehicles. We integrate a lateral-velocity model to further improve the algorithm's accuracy. Moreover, we employ an adaptive Kalman filter for simultaneous state and parameter estimation which further improves accuracy under low-excitation conditions. Fig.~\ref{fig:method} shows the proposed approach in detail from sensor inputs to the system's outputs.

\section{System Description}
\subsection{Notation}
We provide a brief overview of the employed notation. Three coordinate systems are used throughout the paper: the global world frame, ${G}$, the body-fixed coordinate frame, ${B}$, and the camera-fixed coordinate frame, ${C}$. The body-fixed coordinate frame has its origin at the IMU's position and is aligned with the vehicle axes (front, left, up). The global world frame is assumed to be flat, ignoring the Earth's curvature. Moreover, we ignore the rotation of the Earth since automotive-grade IMUs cannot measure it anyway.

Vectors are denoted as lower-case bold letters, e.g. $\mathbf{x}$, scalars as non-bold lower-case letters, e.g. $v$, and matrices as upper-case bold letters, e.g. $\textbf{F}$. Hamilton quaternions are used to represent attitude, e.g. ${}^{G}\mathbf{q}_{B}$ describing the attitude of frame ${B}$ relative to frame ${G}$. The associated rotation matrix is ${}^{G}\mathbf{R}_{B}$, which is part of the special orthogonal group $SO(3)$ \cite{Sola.2017}. A similar notation is used for vectors, e.g. where ${}^{G}\mathbf{p}_{B}$ represents the position vector of frame ${B}$ to frame ${G}$, expressed in frame ${G}$. For ease of readability we define the following simplified notations for frequently used expressions:
\begin{itemize}
\item $\mathbf{p}_{B} := {}^{G}\mathbf{p}_{B}:$ position in the world frame,
\item $\mathbf{v}_{B} := {}^{B}\mathbf{v}_{B}:$ velocity in the body-fixed frame,
\item $\mathbf{q}_{B} := {}^{G}\mathbf{q}_{B}:$ attitude (body-fixed to world frame),
\item $\boldsymbol{\omega}_{B} := {}^{B}\boldsymbol{\omega}_{B}:$ angular rate in the body-fixed frame,
\item $\boldsymbol{a}_{B} := {}^{B}\boldsymbol{a}_{B}:$ acceleration in the body-fixed frame.
\end{itemize}

\subsection{Motion Model}
The motion model describes the vehicle's body-fixed velocity $\boldsymbol{v}_{B}$, its attitude $\mathbf{q}_{B}$ and the position $\boldsymbol{p}_{B}$ using inertial measurements as inputs. The derivatives of these states are expressed as \cite{Sola.2017}
\begin{align}
\label{eq:vmo_1}
\dot{\mathbf{v}}_{B} &= \mathbf{a}_{B}+\mathbf{R}(\mathbf{q}_{B})^\top {}^{G}\mathbf{g}-[\boldsymbol{\omega}_{B}\times] {\mathbf{v}}_{B},\\
\dot{\mathbf{q}}_{B} &= \frac{1}{2}\mathbf{q}_{B}\bullet\begin{bmatrix}
0\\ \boldsymbol{\omega}_{B} 
\end{bmatrix},\label{eq:vmo_2} \\
\dot{\mathbf{p}}_{B} &= \mathbf{R}(\mathbf{q}_{B}){\mathbf{v}}_{B}, \label{eq:vmo_3}
\end{align}
with the gravity vector ${}^{G}\mathbf{g}$, the direction cosine matrix $\mathbf{R}(\mathbf{q}_{B})={}^{G}\mathbf{R}_{B}$, the acceleration $\mathbf{a}_{I}$, the angular rates $\boldsymbol{\omega}_{I}$, where the symbol $\{\bullet\}$ indicates quaternion multiplication, and where $[\boldsymbol{\omega}_{I}\times]$ is a skew-symmetric matrix.
The dynamic state vector is
\begin{align}
\boldsymbol{x}_{B} &= \begin{bmatrix}
{\mathbf{v}}_{B}^\top  & \mathbf{q}_{B}^\top &{\mathbf{p}}_{B}^\top \end{bmatrix}^\top. 
\end{align}
We extract the following Jacobians with respect to the IMU states:
\begin{align}
\dfrac{\partial\, \dot{\mathbf{v}}_{B}}{\partial\, \mathbf{v}_{B}}&=-[{\boldsymbol{\omega}_{B}}\times],\quad
\dfrac{\partial\, \dot{\mathbf{v}}_{B}}{\partial\, \mathbf{q}_{B}}={}^{B}\mathbf{R}_{G}[ {}^{G}\mathbf{g}\times],\\
\dfrac{\partial\, \dot{\mathbf{p}}_{B}}{\partial\, \mathbf{v}_{B}} &= {}^{G}\mathbf{R}_{B}, \quad \dfrac{\partial\, \dot{\mathbf{p}}_{B}}{\partial\, \mathbf{q}_{B}}=-[{}^{G}\mathbf{R}_{B}\mathbf{v}_{B}\times].
\end{align}

\subsection{Relative Feature Motion}
The stationary features move relative to the camera sensor at a rate that is entirely dependent on the vehicle's ego motion. Bloesch et al. \cite{Bloesch.2017} propose to parameterize these features on the unit sphere $S^2$. This separates depth and bearing which is especially relevant in vision-based approaches for sensors that cannot directly measure depth. Moreover, it was shown that using the inverse depth instead of the depth directly within the filter improves consistency \cite{Civera.2008}. The bearing vector of a feature $j$ with bearing
\begin{align}
\mathbf{q}_{f_j}:={}^{C}\mathbf{q}_{f_j}
\end{align}
is calculated as \cite{Bloesch.2017}
\begin{align}
\boldsymbol{p}_{f_j}:={}^{C}\boldsymbol{p}_{f_j} = \mathbf{R}\left(\mathbf{q}_{f_j}\right)\boldsymbol{e}_1 = {}^{C}\mathbf{R}_{f_j} \boldsymbol{e}_1.
\end{align}
The matrix ${}^{C}\mathbf{R}_{f_j}$ rotates the basis vector $\mathbf{e}_1$ from the feature direction into the radar sensor coordinate system and $\mathbf{e}_{1/2/3}\in\mathbb{R}^3$ describe the basis vectors of an orthonormal coordinate system.
We also define the projection matrix:
\begin{align}
\mathbf{N}\left(\mathbf{q}_{f_j}\right) = {}^{C}\mathbf{R}_{f_j}\begin{bmatrix}\boldsymbol{e}_2&\boldsymbol{e}_3\end{bmatrix} = \mathbf{N}_\mathbf{q} \in \mathbb{R}^{3\times 2}.
\end{align}
This matrix is orthogonal to the bearing vector $\boldsymbol{p}_{f_j}$ and is used to reduce the axis-angle representation to this orthogonal plane \cite{Jackson.2019}. The feature dynamics are given as \cite{Bloesch.2017}\cite{Jackson.2019}
\begin{align}
\label{eq:feature1}
\dot{\mathbf{q}}_{f_j} &=-\mathbf{N}_\mathbf{q}^\top \left(\boldsymbol{\omega}_{C}+\rho_{j}\left[\mathbf{p}_{f_j}\times\right]\boldsymbol{v}_{C}\right),\\
\dot{\rho}_{j} &= \rho_j^2\left(C_{f_j}^s\boldsymbol{e}_1\right)^\top \boldsymbol{v}_{C}, \label{eq:feature2}
\end{align}
with the features bearing $\mathbf{q}_{f_j}$ and inverse depth $\rho_j$. The sensor-fixed velocity and turning rates are calculated using
\begin{align}
\boldsymbol{v}_{C} &= {}^{C}\mathbf{R}_{B} \left(\boldsymbol{v}_{B} + \smat{\boldsymbol{\omega}_{B}}{}^{C}\boldsymbol{p}_{B}\right), \\
\boldsymbol{\omega}_{C} &= {}^{C}\mathbf{R}_{B}\boldsymbol{\omega}_{B},
\end{align}
with the camera extrinsic alignment ${}^{C}\mathbf{R}_{B}$ and translation ${}^{C}\boldsymbol{p}_{B}$.
The dynamic feature state vector is
\begin{align}
\boldsymbol{x}_f = \begin{bmatrix}
\mathbf{q}_{f_1}^{\top }&\rho_{1} & \mathbf{q}_{f_2}^{\top }&\rho_{2} & \cdots &\mathbf{q}_{f_j}^{\top }&\rho_{j}
\end{bmatrix}^\top 
\end{align}
We can derive Jacobians w.r.t. the dynamic and feature state vectors.
With respect to the feature states $\boldsymbol{x}_f$ we obtain
\begin{align}
\begin{split}
\dfrac{\partial\,\dot{\mathbf{q}}_{f_j}}{\partial\,\mathbf{q}_{f_j}} =& \mathbf{N}_\mathbf{q}^\top\left[\left(\boldsymbol{\omega}_{C}+[\mathbf{p}_{f_j}\times]\frac{\mathbf{v}_{C}}{\rho_{j}}\right)\times\right]\mathbf{N}_\mathbf{q}\\
&+ \mathbf{N}_\mathbf{q}^\top\frac{1}{\rho_j}[\mathbf{v}_{C}\times][\mathbf{p}_{f_j}\times]\mathbf{N}_\mathbf{q},
\end{split} \\
\dfrac{\partial\,\dot{\mathbf{q}}_{f_j}}{\partial\, \rho_{j}} &= \mathbf{N}_\mathbf{q}^\top[\mathbf{p}_{f_j}\times]\mathbf{v}_{C}\frac{1}{\rho_{j}^2},\\
\dfrac{\partial\,\dot{\rho}_{j}}{\partial\, \mathbf{q}_{f_j}} &= -\mathbf{v}_{C}^\top[\mathbf{p}_{f_j}\times]\mathbf{N}_\mathbf{q}.
\end{align}
The Jacobians w.r.t. the IMU states $\boldsymbol{x}_{B}$ are
\begin{align}
\dfrac{\partial\,\dot{\mathbf{q}}_{f_j}}{\partial\, \mathbf{v}_{C}} &=-\mathbf{N}_\mathbf{q}^\top\frac{1}{\rho_{j}}[\mathbf{p}_{f_j}\times],\\
\dfrac{\partial\,\dot{\rho}_{j}}{\partial\, \mathbf{v}_{C}} &= -\mathbf{p}_{f_j}^\top.
\end{align}
The depth dynamics provide the Jacobians
\begin{align}
\dfrac{\partial\,\dot{\rho}_{j}}{\partial\, \mathbf{q}_{f_j}} &= -\mathbf{v}_{C}^\top[\mathbf{p}_{f_j}\times]\mathbf{N}_\mathbf{q},\\
\dfrac{\partial\,\dot{\rho}_{j}}{\partial\, \mathbf{v}_{C}} &= -\mathbf{p}_{f_j}^\top.
\end{align}

\subsection{Gyroscope Error Model}
As we want to calibrate the gyroscope, we extend the dynamic system to account for gyroscope errors
\begin{align}
\label{eq:om_err}
{\boldsymbol{\omega}}_{B}^{m} = \begin{bmatrix}
1 & 0 & -s_{\omega_{yx}}\\
0 & 1 & s_{\omega_{xy}}\\
0 & 0 & s_{\omega_{z}}
\end{bmatrix} \boldsymbol{\omega}_{B} + \boldsymbol{b}_\omega,
\end{align}
with the measured angular rates ${\boldsymbol{\omega}}_{B}^{m}$, the gyroscope offsets $\boldsymbol{b}_\omega$, the misalignment parameters $s_{\omega_{yx}}$ and $s_{\omega_{xy}}$, and the yaw-rate scale error $s_{\omega_z}$. In visual-inertial calibration literature additional calibration parameters are typically used, e.g. the complete misalignment matrix. Given the limited excitation of consumer-grade vehicles, error models with additional parameters would not be observable \cite{Yang.2020}. The gyroscope's excitation in consumer-grade vehicles is dominated by the yaw rate $\mathbf{\omega}_{B,z}$. This is why we only choose scale parameters in connection with the yaw rate. Both the roll and pitch rates are excited too little and too infrequently to both allow for proper calibration and to even have a significant impact on localization performance.
We define the parameter vector:
\begin{align}
\boldsymbol{x}_\rho = \begin{bmatrix}
\boldsymbol{b}_\omega^{\,\top }&s_{\omega_z}&s_{\omega_{yx}}&s_{\omega_{xy}}
\end{bmatrix}^\top \in\mathbb{R}^{6}
\end{align}
Using the chain rule we calculate the Jacobians of the dynamics systems w.r.t. the parameter vector
\begin{align}
\dfrac{\partial \dot{\boldsymbol{v}}_{B}}{\partial\boldsymbol{x}_\rho} &=  [{\boldsymbol{v}}_{B}\times] \dfrac{\partial\boldsymbol{\omega}_{B}}{\partial\boldsymbol{x}_\rho},\\
\dfrac{\partial \dot{\mathbf{q}}_{B}}{\partial\boldsymbol{x}_\rho} &= -\dfrac{\partial\boldsymbol{\omega}_{B}}{\partial\boldsymbol{x}_\rho},
\end{align}
where $\frac{\partial\boldsymbol{\omega}_{B}}{\partial\boldsymbol{x}_\rho}$ is the the partial derivative of Eq.~\eqref{eq:om_err} w.r.t. the parameter vector $\boldsymbol{x}_\rho$.
The Jacobians of the relative feature motion w.r.t. the parameter vector are
\begin{align}
\dfrac{\partial\dot{\mathbf{q}}_{f_j}}{\partial\boldsymbol{x}_\rho} &= \left(-\mathbf{N}_\mathbf{q}^\top {}^{C}\mathbf{R}_{B} + \mathbf{N}_\mathbf{q}^\top \rho_j[\boldsymbol{p}_{f_j}\times]{}^{C}\mathbf{R}_{B}[{}^{C}\boldsymbol{p}_{B}\times]\right)\dfrac{\partial\boldsymbol{\omega}_{B}}{\partial\boldsymbol{x}_\rho},\\
\dfrac{\partial\dot{\rho}_{j}}{\partial\boldsymbol{x}_\rho} &=\boldsymbol{p}_{f_j}^{\top}{}^{C}\mathbf{R}_{B}[{}^{C}\boldsymbol{p}_{B}\times]\dfrac{\partial\boldsymbol{\omega}_{B}}{\partial\boldsymbol{x}_\rho}.
\end{align}

\subsection{Camera Measurement}
The camera measurement and feature handling largely follows the proposals of Bloesch et al. \cite{Bloesch.2017}. New features are detected using the FAST feature detector \cite{Rosten.2006} after which we extract an image pyramid at the feature's position. The measurement model follows the Kanade-Lucas-Tomasi (KLT) feature tracker \cite{Lucas.1981, Tomasi.1994}, which updates the filter based on the pixel-intensity error $\Delta I(u,v)$ at position $[u,v]$. This requires the intrinsic camera calibration model that maps the pinhole-camera pixels $[r_x,r_y]$ to the camera pixels
\begin{align}
[u,v] = \text{f}\,(r_x,r_y).
\end{align}
We obtain the measurement Jacobian
\begin{align}
\dfrac{\partial \Delta I({u},{v})}{\partial\mathbf{q}_{f_j}} = \dfrac{\partial \Delta I({u},{v})}{\partial [u,\ v]}\dfrac{\partial [u,\ v]}{\partial[r_x,\ r_y]}\dfrac{\partial [r_x,\ r_y]}{\partial\boldsymbol{p}_{f_j}}\dfrac{\partial\boldsymbol{p}_{f_j}}{\partial \mathbf{q}_{f_j}},
\end{align}
where we apply the chain rule to initially calculate the Jacobian w.r.t. the pixel position ${\partial [u,\ v]}$, the subsequent camera-intrinsics Jacobian w.r.t. the pinhole-camera pixels ${\partial[r_x,\ r_y]}$, then the Jacobian w.r.t. the feature position ${\partial\boldsymbol{p}_{f_j}}$, and finally the Jacobian w.r.t. the feature's bearing ${\partial \mathbf{q}_{f_j}}$.
Both the intrinsic and extrinsic camera-error models are already calibrated in consumer-grade vehicles during an end-of-line calibration process \cite{Ding.2025}.

\subsection{Vehicle Measurement Model}
We now introduce the velocity measurements, focusing on our contribution, the model-based lateral-velocity measurement. We propose using the single-track-model-based lateral-velocity model that depends on one parameter only
\begin{align}
v_{y,r} = -\rho_{sg} a_{B,y} v_{B,x}.
\end{align}
The model consists of the lateral velocity at the rear axis $v_{y,r}$, the lateral acceleration $a_{B,y}$, the velocity $v_{B,x}$, and the side-slip gradient $\rho_{sg}$. The side-slip gradient is a vehicle-specific parameter that can be estimated during vehicle testing \cite{Diener.2025}. This proposed velocity model is well-established in vehicle-dynamics literature \cite{Diener.2024} and is
especially suitable for normal driving ($v>10$ m/s, $a_{B,y}<4$ m/s$^2$) \cite{Marco.2020}.

Xiong et al. \cite{Xiong.2022} also integrated vehicle-dynamics constraints into their SLAM algorithm. However, they used many vehicle-specific parameters that make their approach unsuitable for consumer-grade applications. Additionally, they used the single-track-model-based yaw-rate measurement \cite{Xiong.2022} that introduces even more uncertain parameters and a sensor (wheel-steering angle) that requires calibration. 

The longitudinal velocity is measured using wheel encoders. For the vertical velocity we make a zero-mean assumption. We obtain the vehicle-velocity measurement vector
\begin{align}
\mathbf{y}_v^\text{m} = \begin{bmatrix}
v_{B,x}^\text{m} \\ -\rho_{sg} a_{B,y} v_{B,x} \\ 0
\end{bmatrix}.
\end{align}

\section{Localization Filter}
We propose to use an adaptive Kalman filter \cite{Zhang.2022} for simultaneous state and parameter estimation. The standard approach for combined state and parameter estimation is augmentation of the system state \cite{Yang.2024}. The main issue is that this can lead to filter divergence when the system is insufficiently excited \cite{Marco.2022}. The adaptive Kalman filter separates state and parameter estimation, which stabilizes the filter during situations with limited excitation.
Our linearized continuous-time system is
\begin{align}
\dot{\boldsymbol{x}} = \mathbf{F}\,\boldsymbol{x} &= \left[ \begin{array}{c|ccc}
\mathbf{F}_{B} & \multicolumn{3}{|c}{\mathbf{0}_{9\times 6+3j}}\\
\hline
\mathbf{F}_{B_1} &\mathbf{F}_{f_1} & &\\
\vdots & & \ddots &\\
\mathbf{F}_{B_j} & & &\mathbf{F}_{f_j}
\end{array}
\right]
\begin{bmatrix}
\boldsymbol{x}_{B}\\\boldsymbol{x}_f
\end{bmatrix} +\mathbf{\Psi}\boldsymbol{x}_\rho,\\
\mathbf{y} = \mathbf{H}\,\boldsymbol{x} &= 
\left[ \begin{array}{c|ccc}
\mathbf{H}_{v} & \multicolumn{3}{|c}{\mathbf{0}_{3\times 3j}}\\
\hline
 &\mathbf{H}_{f_1} & &\\
\mathbf{0}_{2j\times 9} & & \ddots &\\
 & & &\mathbf{H}_{f_j}
\end{array}
\right]
\begin{bmatrix}
\boldsymbol{x}_B\\\boldsymbol{x}_f
\end{bmatrix}.
\end{align}
The system features the IMU state vector $\boldsymbol{x}_{B}$, the feature vector $\boldsymbol{x}_f$, its Jacobians
\begin{align}
\mathbf{F}_{B} &= \left[
\begin{array}{ccc}
\frac{\partial \dot{\mathbf{v}}_{B}}{\partial\mathbf{v}_{B}} & \frac{\partial \dot{\mathbf{v}}_{B}}{\partial\mathbf{q}_{B}} & \mathbf{0}_{3}\\
\mathbf{0}_{3} & \mathbf{0}_{3} & \mathbf{0}_{3} \\
\frac{\partial \dot{\mathbf{p}}_{B}}{\partial\mathbf{v}_{B}} & \frac{\partial \dot{\mathbf{p}}_{B}}{\partial\mathbf{q}_{B}}& \mathbf{0}_{3}
\end{array}
\right],\\
\mathbf{F}_{B_j} &= \begin{bmatrix}
\frac{\partial\dot{\mathbf{q}}_{f_j}}{\partial \mathbf{v}_{B}} & \mathbf{0}_{2\times 3}& \mathbf{0}_{2\times 3}\\
\frac{\partial\dot{\rho}_{j}}{\partial \mathbf{v}_{B}}& \frac{\partial\dot{\rho}_{j}}{\partial \mathbf{q}_{f_j}} & \mathbf{0}_{1\times 3}
\end{bmatrix},\
\mathbf{F}_{f_j} = \begin{bmatrix}
\frac{\partial\dot{\mathbf{q}}_{f_j}}{\partial\mathbf{q}_{f_j}} & \frac{\partial\dot{\mathbf{q}}_{f_j}}{\partial \rho_{j}}
\end{bmatrix},
\end{align}
Moreover, the Jacobian w.r.t. the parameter state is
\begin{align}
\mathbf{\Psi} = \begin{bmatrix}
\frac{\partial \dot{\boldsymbol{v}}_{B}}{\partial\boldsymbol{x}_\rho}^\top  & \frac{\partial \dot{\mathbf{q}}_{B}}{\partial\boldsymbol{x}_\rho}^\top & \mathbf{0}_{4\times 3}&\frac{\partial\dot{\mathbf{q}}_{f_j}}{\partial\boldsymbol{x}_\rho}^\top  &\frac{\partial\dot{\rho}_{j}}{\partial\boldsymbol{x}_\rho}^\top 
\end{bmatrix}^\top ,
\end{align}
and the measurement matrices are
\begin{align}
\mathbf{H}_v &=\begin{bmatrix}
1 & 0 & 0\\
{\rho}_{sg}a_{B,y} & 1 & 0\\
0 & 0 & 1
\end{bmatrix},\\ \mathbf{H}_{f_j} &= \begin{bmatrix}
\dfrac{\partial \Delta I({u},{v})}{\partial\mathbf{q}_{f_j}^s} & \mathbf{0}_{2\times 1}
\end{bmatrix}.
\end{align}

For the adaptive filter we use a variation of \cite{Marco.2020} and \cite{Zhang.2018}.  The system's state $\boldsymbol{x}_{k|k}$ at $t=t_k$ is propagated using numerical integration of the continuous-time differential equations describing the IMU motion (Eqs.~\eqref{eq:vmo_1}-\eqref{eq:vmo_3}) and feature motion (Eqs.~\eqref{eq:feature1}-\eqref{eq:feature2}), resulting in the propagated state estimate $\boldsymbol{x}_{k+1|k}$. However, the covariance $\mathbf{P}_{k|k}$ is propagated using the linearized discrete-time system
\begin{align}
\dot{\boldsymbol{x}}_{k+1|k} &= \boldsymbol{\Phi}_k\,\boldsymbol{x}_{k|k} + \mathbf{\Psi}\boldsymbol{x}_\rho + \boldsymbol{w}_k ,\\
\mathbf{y}_k &= \mathbf{H}\,\mathbf{x}_{k|k} + \boldsymbol{v}_k,
\end{align}
where $\boldsymbol{w}_k$ is the process noise with covariance matrix $\mathbf{Q}_k$, where $\boldsymbol{v}_k$ is the measurement noise with covariance matrix $\mathbf{R}_k$. We obtain the discrete-time state transition matrix $\boldsymbol{\Phi}_k$ through Euler integration of $\mathbf{F}(t)$
The state's covariance matrix $\mathbf{P}_{k|k}$ can then be propagated from $t_k$ to $t_{k+1}$ using
\begin{align}
\mathbf{P}_{k+1|k} = \boldsymbol{\Phi}_k \mathbf{P}_{k|k} \boldsymbol{\Phi}_k^\top + \mathbf{Q}_k.
\end{align}
The time update consists of the regular Kalman-filter innovation and the adaptive parameter estimation that is similar to a recursive-least-squares (RLS) algorithm \cite{Zhang.2018}. We thus first perform the Kalman-filter innovation
\begin{align}
\mathbf{\Sigma}_k &= \mathbf{H}\mathbf{P}_{k+1|k}\mathbf{H}^\top+\mathbf{R}_k,\\
\mathbf{K}_k &= \mathbf{P}_{k+1|k} \mathbf{H}^\top\mathbf{\Sigma}_k^{-1},\\
\mathbf{P}_{k+1|k+1} &= \left(\mathbf{I}-\mathbf{K}_k\mathbf{C}\right)\mathbf{P}_{k+1|k},
\end{align}
with the identity matrix $\mathbf{I}$, the innovation covariance $\mathbf{\Sigma}_k$ and the Kalman-Gain $\mathbf{K}_k$. The adaptive part is similar to an RLS algorithm. We first obtain the matrix of regressors
\begin{align}
\mathbf{\Omega}_k = \mathbf{H}\mathbf{\Phi}_k\mathbf{\Upsilon}_{k}+\mathbf{H}\mathbf{\Psi}_k,
\end{align}
with use of the auxiliary variable
\begin{align}
\mathbf{\Upsilon}_{k+1} = \left(\mathbf{I}-\mathbf{K}_k\mathbf{H}\right)\mathbf{\Phi}_k\mathbf{\Upsilon}_{k}+\left(\mathbf{I}-\mathbf{K}_k\mathbf{H}\right)\mathbf{\Psi}_k.
\end{align}
This, together with the parameter covariance $\mathbf{S}_k$ and the forgetting factor $\lambda$, we use to perform the RLS innovation
\begin{align}
\mathbf{\Lambda}_k &= \left(\mathbf{\Sigma}_k+\mathbf{\Omega}_k\mathbf{S}_{k}\mathbf{\Omega}_k^\top\right)^{-1},\\ 
\mathbf{\Gamma}_k &=\mathbf{S}_{k}\mathbf{\Omega}_k^\top\Lambda_k, \\
\mathbf{S}_{k+1} &= \frac{1}{\lambda}\left(\mathbf{S}_{k}\mathbf{\Omega}_k^\top\mathbf{\Lambda}_k\mathbf{\Omega}_k\right)\mathbf{S}_{k}.
\end{align}
Finally, the parameter and state vectors are updated
\begin{align}
\boldsymbol{x}_{\rho,k+1} &=\boldsymbol{x}_{\rho,k}+\mathbf{\Gamma}_k\tilde{\mathbf{y}}_k,\\
\boldsymbol{x}_{k+1|k+1} &= \boldsymbol{x}_{k+1|k}+ \mathbf{K}_k\tilde{\mathbf{y}}_k +\mathbf{\Upsilon}_k\mathbf{\Gamma}_k\tilde{\mathbf{y}}_k,
\end{align}
using the measurement residual
\begin{align}
\tilde{\mathbf{y}}_k = \mathbf{y}_k-{\mathbf{y}}^\text{m}_k.
\end{align}


\section{Experimental Results}
Our experimental evaluation focuses on gyroscope calibration accuracy and overall localization performance. We evaluate using both proprietary and public data. We collected the proprietary dataset on a consumer-grade vehicle using on-board sensors, i.e. the front-facing camera, the automotive-grade IMU, and the wheel encoders. Moreover, we equipped the vehicle with a ground-truth measurement system to obtain an accurate reference. We also use the public KAIST dataset \cite{Jeong.2019} also offers a front-facing camera, a high-precision IMU, and a ground-truth trajectory.

\begin{figure}[b]
\centering
\includegraphics[width=\linewidth]{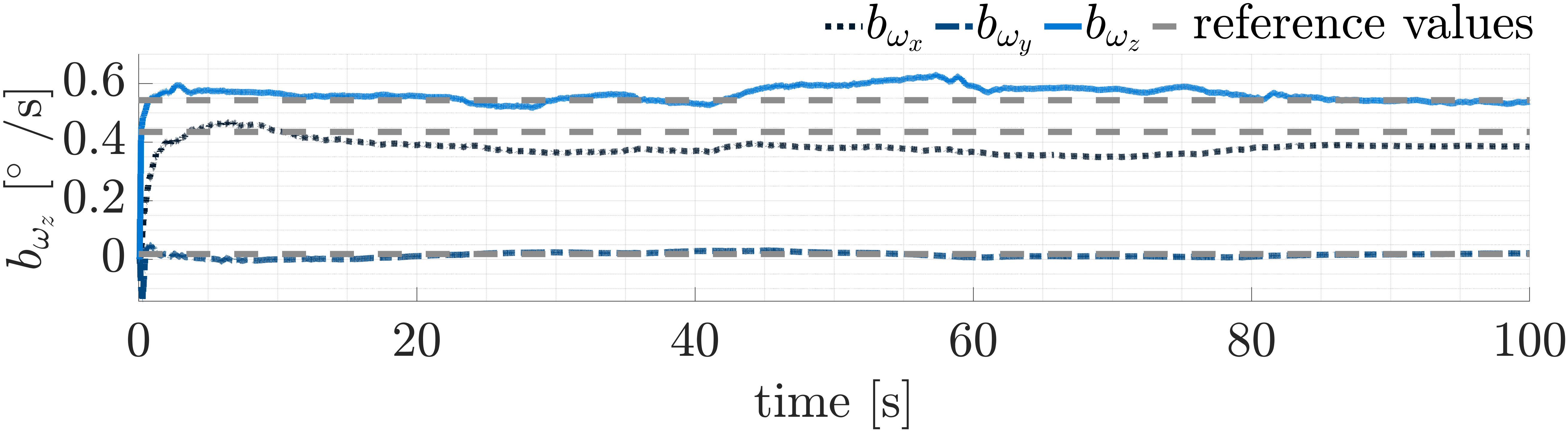}
\caption{Estimated offsets $\mathbf{b}_{\omega}$ over time with their ground-truth values.}
\label{fig:ex1}
\end{figure}

\begin{figure}[t]
\centering
\includegraphics[width=\linewidth]{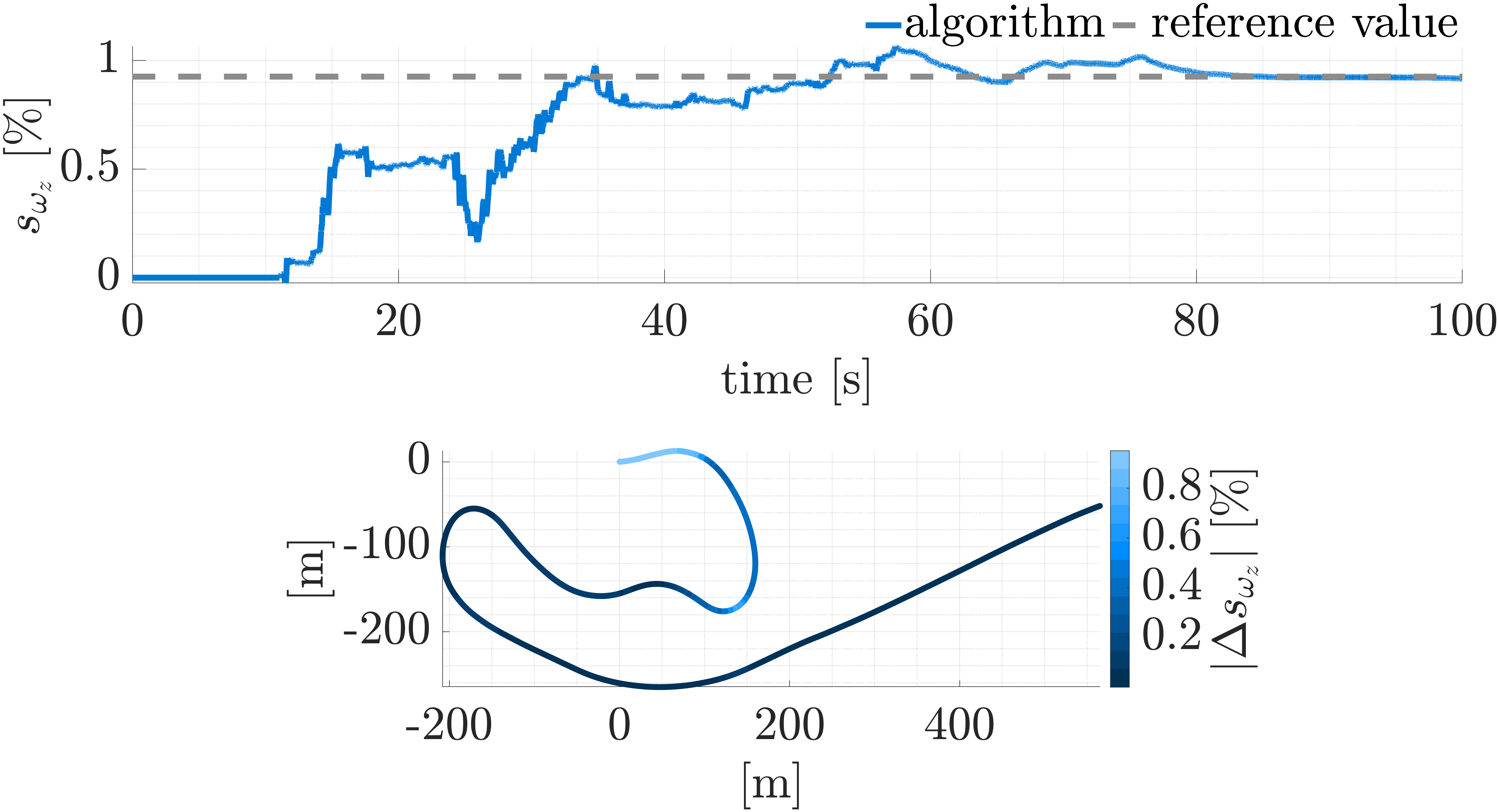}
\caption{Estimated yaw-rate scale error $s_{\omega_z}$ over time (above). Lower plot depicts the deviation from the ground-truth value over the GNSS position.}
\label{fig:ex2}
\end{figure}

\begin{figure}[b]
\centering
\includegraphics[width=\linewidth]{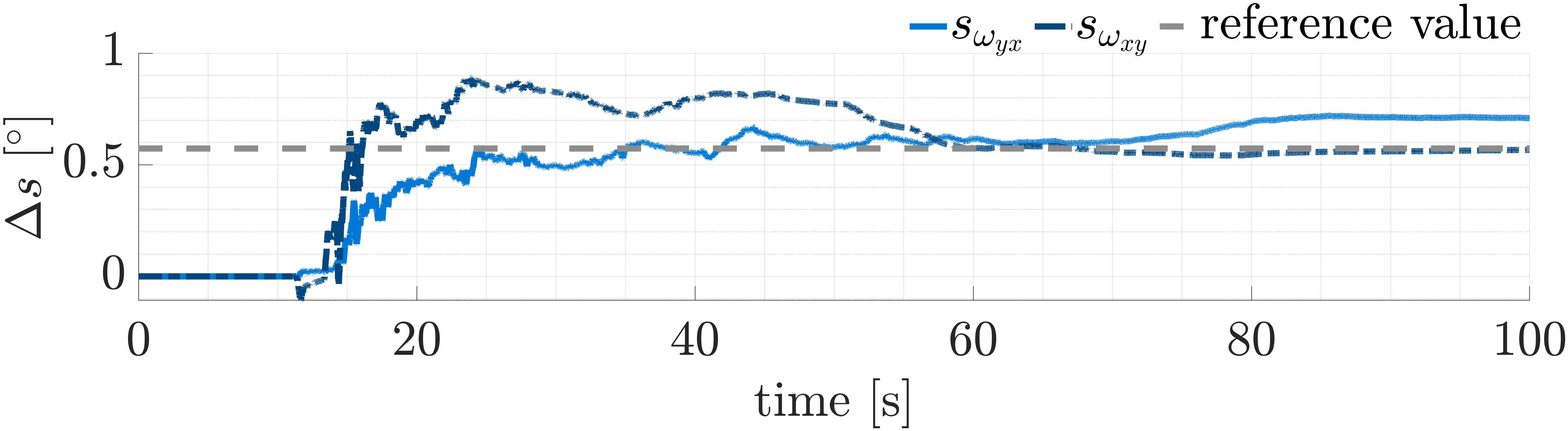}
\caption{Estimated misalignment parameters $s_{\omega_{yx}}$ and $s_{\omega_{xy}}$ and their reference value.}
\label{fig:ex3}
\end{figure}

%
%

\subsection{Gyroscope Calibration}
We initially analyze gyroscope calibration performance of our approach on the proprietary dataset with the automotive-grade IMU. Our calibration model consists of six calibration parameters, i.e. the gyroscope offsets $\mathbf{b}_{omega}$, the yaw-rate scale error $s_{\omega_z}$, and the yaw-rate affected misalignments $s_{\omega_{yx}}$ and $s_{\omega_{xy}}$. The ground-truth values of the calibration parameters were obtained using a high-precision reference IMU. Fig.~\ref{fig:ex1} depicts the online calibration of the gyroscope offsets $\mathbf{b}_\omega$ using our approach and the accompanying reference values. All offsets quickly converge towards their true values within a few seconds. The final calibration results are within 0.05$^\circ$/s of their true values.

Next, we consider the calibration of the yaw-rate scale error $s_{\omega_z}$. The calibration of this particular parameter requires yaw excitation, which is why convergence takes more time. Fig.~\ref{fig:ex2} depicts the online calibration of the yaw-rate scale error and additionally shows the convergence behavior over the GNSS position of the vehicle. After the second turn, the parameter is estimated within 0.1\% of the true value. This shows that at least some form of yaw excitation is necessary to achieve convergence. However, these excitation requirements are not restrictive for consumer-grade vehicles.

Last, we evaluate the calibration performance of the two misalignment parameters $s_{\omega_{yx}}$ and $s_{\omega_{xy}}$. Fig.~\ref{fig:ex3} shows the online calibration of the two parameters over time. Once the vehicle exhibits yaw excitation, our algorithm quickly converges toward the reference value. The misalignment is then calibrated within 0.1$^\circ$ of this reference value.

\subsection{Localization Performance}
We demonstrated that our proposed algorithm enables the online calibration of the gyroscope error model. In the following, we evaluate its localization performance and compare it against wheel-IMU odometry. By wheel-IMU odometry, we refer to our localization algorithm without feature tracking, i.e., it solely relies on the IMU motion model combined with the vehicle’s velocity measurements. In this configuration, the IMU is only offset-calibrated, while the yaw-rate scale error and misalignment parameters remain uncalibrated.

\renewcommand{\arraystretch}{1.3}
\begin{table}[b]
\centering
\begin{threeparttable}
\caption{Localization performance of various algorithm variants and the full solution.}
\label{tab:localization}
\begin{tabular}{wl{3.9cm}|wc{1.2cm}wc{1.2cm}wc{0.6cm}}
\hline
& {63rd \%-ile} & {95th  \%-ile} & {max.} \\
\hline
wheel-IMU odom. (uncalibrated) & 1.09 \% & 4.51 \% & 8.26 \% \\
solution w/o gyro calibration & \underline{0.69 \%} & 5.88 \% & 9.42 \% \\
solution w/o lateral-velocity model & {1.77 \%} & 5.41 \% & 12.68 \% \\
\hline
\textbf{full solution} & 0.80 \% & \underline{3.10 \%} & \underline{6.32 \%}\\
wheel-IMU odom. \textbf{(calibrated)} & \textbf{0.38 \%} & \textbf{1.78 \%} & \textbf{5.85 \%}
\end{tabular}
\begin{tablenotes}
\item Note: Values in \textbf{bold} indicate the best results and \underline{underlined} values indicate the second-best results.
\end{tablenotes}
\end{threeparttable}
\end{table}

\begin{figure*}[t]
\centering
\includegraphics[width=0.9\textwidth]{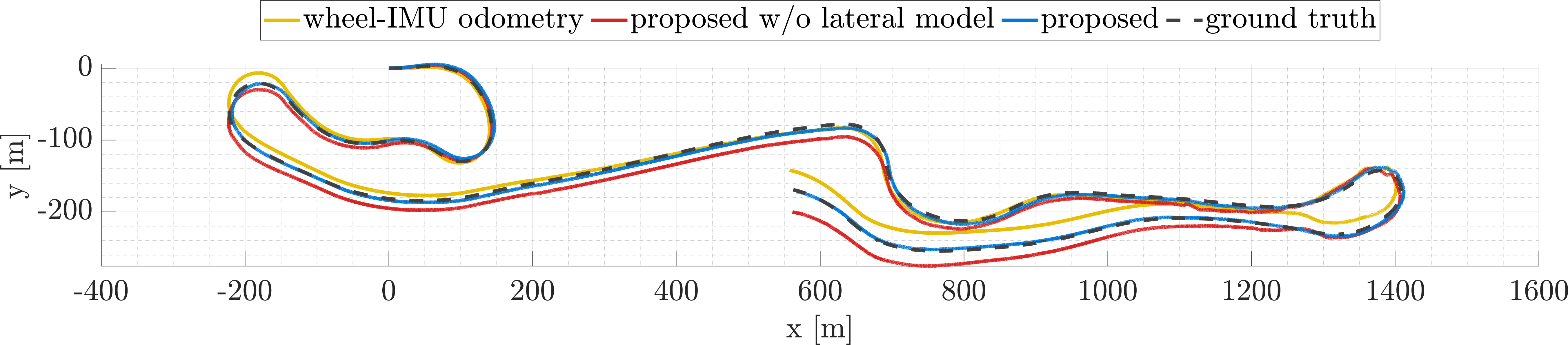}
\caption{Localization Results on proprietary dataset: ground truth (dashed), wheel-IMU odometry (yellow,) proposed solution without lateral-velocity model (red), proposed solution without calibration (green), proposed solution (blue). }
\label{fig:local}
\end{figure*}

Fig.~\ref{fig:local} illustrates the estimated trajectories of different variants of our algorithm along with the ground-truth trajectory. We observe that our full solution most closely follows the ground truth, whereas the pure wheel-IMU odometry shows a visible deviation. Similarly, removing the lateral-velocity model, one of our key contributions, results in a clear drift from the reference trajectory.

To substantiate these qualitative observations, we perform a statistical evaluation of the localization accuracy using the relative pose error (RPE). The RPE is defined following the KITTI metric~\cite{Geiger.2013} over a path length of 100~m, such that an RPE of 1\% corresponds to a localization error of 1~m over these intervals. The evaluated sequence spans a total distance of approximately 18~km. For statistical comparison, we consider the 63rd percentile (performance threshold), the 95th percentile (consistency threshold), and the maximum RPE value. The corresponding results are summarized in Table~\ref{tab:localization}.

Our full solution achieves 0.8\% accuracy most of the time and a consistent accuracy of 3.1\%. Compared to the uncalibrated wheel-IMU odometry (offset calibration only), our method consistently achieves lower errors across all statistical thresholds. The second observation arises from our ablation studies: removing the lateral-velocity model nearly doubles the RPE in all categories, underscoring its importance for accurate localization. When running our algorithm without IMU calibration the overall accuracy degrades. We attribute the marginally lower RPE at the 63rd percentile to additional noise introduced during the gyroscope calibration process.

Finally, we compare our approach with calibrated wheel-IMU odometry, where the calibration parameters are obtained using our online-estimation algorithm. This configuration yields the best overall performance across all metrics. These results are in line with the findings of Gentil~et~al.~\cite{Gentil.2025}, confirming that calibrated wheel-IMU odometry enables highly accurate localization. Moreover, the results demonstrate that proprioceptive localization benefits from integrating visual SLAM through gyroscope calibration. Consequently, even if visual SLAM is temporarily unavailable, the resulting performance improvements remain consistent.

\subsection{Public Dataset}


\renewcommand{\arraystretch}{1.3}
\begin{table}[b]
\centering
\begin{threeparttable}
\caption{Comparison to state-of-the-art visual SLAM approaches.}
\label{tab:public}
\begin{tabular}{l|wc{1.4cm}}
\hline
\textbf{RMSe}& \textbf{translation}\\ 
\hline
VIO \cite{Lee.2021} & {66.31 m} \\
Wheel-VIO \cite{Lee.2020} & 42.74 m \\
Lidar-VIO \cite{Lee.2021} & 26.30 m\\
Zhao et al. \cite{Zhao.2023} & \underline{10.60 m} \\
RLD-SLAM \cite{Zheng.2024} & 34.38 m\\ 
\hline
our solution w/o lateral-velocity model & 11.49 m\\
\textbf{our solution w/o gyro calibration} & \textbf{9.88 m} 
\end{tabular}
\begin{tablenotes}
\item Note: Values in \textbf{bold} indicate the best results and \underline{underlined} values indicate the second-best results.
\end{tablenotes}
\end{threeparttable}
\end{table}

We additionally use the KAIST \textit{urban39} dataset~\cite{Jeong.2019} to benchmark our proposed localization algorithm against state-of-the-art SLAM approaches. Specifically, we compare against pure VIO \cite{Lee.2021}, wheel-VIO \cite{Lee.2020} (tightly coupled wheel-encoder aided VIO), lidar-VIO \cite{Lee.2021} (VIO aided by laser scanner measurements), RLD-SLAM \cite{Zheng.2024} (VIO with loop-closure capability), and the method of Zhao~et~al.~\cite{Zhao.2023}, which incorporates vehicle-motion constraints. Our algorithm is evaluated without online calibration since the IMU provided by the dataset is already calibrated and of high precision.

We evaluate the root-mean-square (RMS) errors of both the trajectory position and rotation over the entire sequence. The corresponding results for all methods are summarized in Table~\ref{tab:public}.
Our proposed solution achieves the highest localization accuracy, with an RMS position error of 9.88~m, surpassing the competing approaches by several meters. The method of Zhao~et~al.~\cite{Zhao.2023}, which also exploits vehicle-motion constraints, achieves the second-best performance, which highlights the importance of incorporating vehicle-dynamics models into visual-inertial localization. Furthermore, when our lateral-velocity model is omitted, the localization accuracy degrades, confirming that this model represents a key contribution to the overall performance of our approach.

\section{Conclusion}
In this work, we integrated visual SLAM into a proprioceptive localization framework and performed online gyroscope calibration. The proposed approach introduces a lateral-velocity model that enhances localization accuracy and an adaptive Kalman filter that enables stable joint state and parameter estimation. We demonstrated that the six-parameter gyroscope error model can be calibrated online under realistic driving conditions. The integration of visual SLAM into proprioceptive localization was shown to improve overall accuracy when gyroscope calibration is performed. Furthermore, the performance improvements persist even when visual SLAM is temporarily unavailable, as the estimated calibration parameters maintain their effect. Benchmarking on a public dataset confirmed that our approach achieves superior localization accuracy compared to state-of-the-art SLAM algorithms.

Future work could focus on extending the gyroscope error model, investigating accelerometer calibration, and exploring the application of the proposed framework to low-speed and parking scenarios.

\bibliographystyle{IEEEtran}
\bibliography{bib/literatur}

\end{document}